\pdfoutput=1

\documentclass[11pt]{article}
\usepackage[preprint]{coling}

\usepackage{times}
\usepackage{latexsym}
\usepackage{tabularx}
\usepackage{amsmath}
\usepackage{hyperref}
\usepackage[T1]{fontenc}
\usepackage[utf8]{inputenc}
\usepackage{microtype}
\usepackage{inconsolata}
\usepackage{graphicx}
\usepackage{caption} 
\usepackage{booktabs} 
\usepackage{array} 

\title{1-800-SHARED-TASKS at RegNLP: Lexical Reranking of Semantic Retrieval (LeSeR) for Regulatory Question Answering}

\author{Jebish Purbey \\
  Pulchowk Campus, IoE \\
  \texttt{jebishpurbey@gmail.com} \\
  \And Drishti Sharma \\
  Cohere For AI Community \\
  \texttt{drishtishrma@gmail.com} \\ 
  \And Siddhant Gupta *\\
  IIT Roorkee\\
  \texttt{siddhant\_g@me.iitr.ac.in} \\ \\
  \AND Khawaja Murad *\\
  NUST, Pakistan \\
  \texttt{khawajamurad@outlook.com} \\ \\
  \And Siddartha Pullakhandam \\
  University of Wisconsin\\
  \texttt{pullakh2@uwm.edu} \\ \\
  \And Ram Mohan Rao Kadiyala \\
  University of Maryland \\
  \texttt{rkadiyal@umd.edu} \\ }

\begin{document}
\maketitle
\begin{abstract}
This paper presents the system description of our entry for the COLING 2025 RegNLP RIRAG (Regulatory Information Retrieval and Answer Generation) challenge, focusing on leveraging advanced information retrieval and answer generation techniques in regulatory domains. We experimented with a combination of embedding models, including Stella, BGE, CDE, and Mpnet, and leveraged fine-tuning and reranking for retrieving relevant documents in top ranks.
We utilized a novel approach, LeSeR, which achieved competitive results with a recall@10 of \(0.8201\) and map@10 of \(0.6655\) for retrievals. This work highlights the transformative potential of natural language processing techniques in regulatory applications, offering insights into their capabilities for implementing a retrieval augmented generation system while identifying areas for future improvement in robustness and domain adaptation.
\end{abstract}

\makeatletter
\def\blfootnote{\gdef\@thefnmark{}\@footnotetext}
\makeatother
\blfootnote{* equal contribution}

\section{Introduction}

Regulatory documents pose significant challenges for organizations seeking to ensure compliance owing to their complexity and ever-changing nature. It is important for organizations to adhere to regulations to maintain legal compliance. With the recent advances in Natural Language Processing (NLP), there is an opportunity to tackle these issues and automate the process of information retrieval, regulatory comparisons, and compliance verifications. Regulatory NLP (RegNLP) focuses on improving access to and understanding of regulatory rules and obligations by leveraging NLP techniques. Within RegNLP, usage of language models for retrieval of regulatory guidelines for Question Answering (Q/A) has shown great potential \cite{9920030}.\\
In light of this, this paper focuses on our submission to the COLING 2025 Regulatory Information Retrieval and Answer Generation (RIRAG) challenge, involving two key tasks: retrieving top-k relevant passages for the given set of queries and using the relevant passages to formulate answers with language models. Our approach enhances the capabilities of semantic retrievals for RIRAG by fine-tuning an embedding model on positive data pairs and reranking it using lexical retrieval techniques.\\
LeSeR (Lexical reranking of Semantic Retrieval) is a novel hybrid approach that combines dense semantic retrieval with classical lexical reranking for enhanced retrieval performance. It leverages dense embeddings fine-tuned on query-passage pairs and integrates BM25 \cite{bm25} scores to improve ranking precision. This dual approach enables robust retrieval in complex regulatory domains, outperforming both pure lexical and semantic models. We test a multitude of open-source models and select the best model for LeSeR. Our work contributes to developing specialized retrieval systems for Q/A in regulatory domains.

\begin{table*}[h]
\centering
\begin{tabular}{lccccccc}
\toprule
Split & Questions & 1 Passage & 2 Passages & 3 Passages & 4 Passages & 5 Passages & 6 Passages \\ \midrule
Train & 22,295    & 16,946    & 4,016      & 975        & 202        & 100        & 56         \\
Test  & 2,786     & 2,126     & 506        & 105        & 36         & 9          & 4          \\
Dev   & 2,888     & 2,215     & 514        & 116        & 30         & 12         & 1          \\ \bottomrule
\end{tabular}
\caption{Distribution of passages per question across train, test, and development splits}
\label{tab:dataset-splits}
\end{table*}

\begin{figure*}[h]
    \centering
  \includegraphics[scale=0.47]{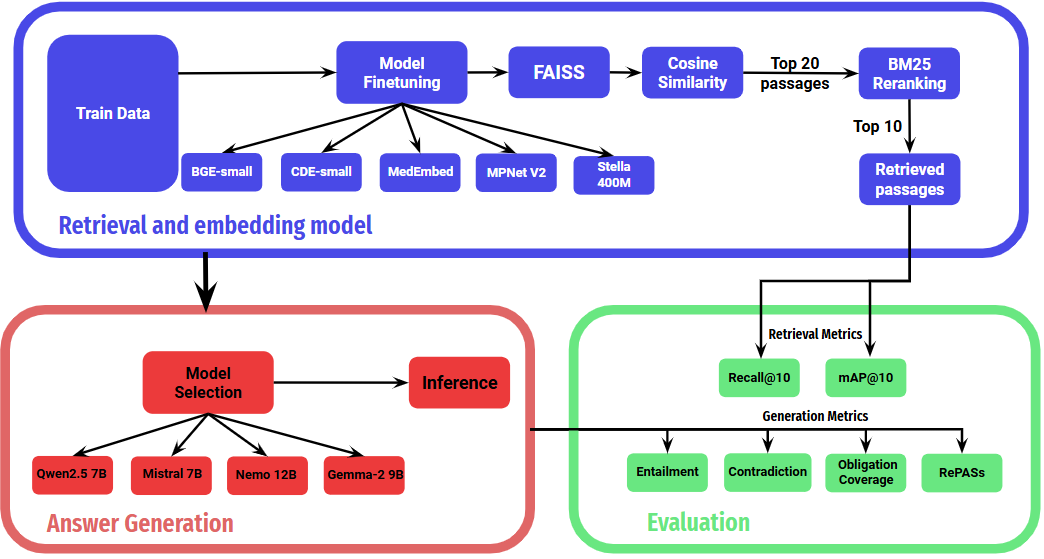}
  \caption{System design workflow}
  \label{fig:system}
\end{figure*}
\section{Dataset \& Task}

The RIRAG task aims to enhance the efficiency and accuracy of compliance-related tasks by addressing two critical subtasks in RegNLP: passage retrieval (Sub-task 1) and answer generation (Sub-task 2). The first Sub-task (1) is to identify and extract the most relevant passages, specifically obligations and related rules, from ADGM regulations and guidance documents. Building on this, the second Sub-task (2) focuses on the ability to generate clear and concise information from varying sources to fully address the compliance and obligation requirement of the query.
RIRAG utilizes the ObliQA Dataset \cite{gokhan2024regnlpactionfacilitatingcompliance} which is a  RegNLP resource built from the Abu Dhabi Global Markets (ADGM) regulations. The dataset incorporates comprehensive and meticulously organized documents, preserving the intricate structure and terminology characteristics of legal and regulatory texts. The dataset includes 22,295 training, 2,888 development, and 2,786 test examples. Each instance consists of a natural language question, relevant regulatory passages annotated with DocumentID and PassageID, and contextual group identifiers. ObliQA is a multi-retrieval dataset and its distribution is given in Table \ref{tab:dataset-splits}.

\section{Methodology}

For the passage retrieval task, our approach leverages a combination of dense and sparse retrieval methods to maximize the relevance and diversity of the retrieved passages as shown in Figure \ref{fig:system}. Hybrid retrieval approaches combine the strengths of semantic and lexical techniques to enhance retrieval quality. In these methods, semantic embeddings are often used for relevance matching, while lexical models ensure precision by addressing nuances like exact term matching and linguistic variation. Such approaches generally merge semantic and lexical scores during retrieval, rather than isolating the two stages. We propose LeSeR (Lexical-Semantic Retrieval), a novel take on hybrid retrieval that uniquely decouples these phases. Semantic embeddings retrieve high-recall candidates, which are then reranked lexically for precision. This strict modularity ensures optimal performance tailored to the challenges of regulatory information retrieval.\\
We utilize a dense vector-based search mechanism using the FAISS library \cite{douze2024faisslibrary}, with embeddings generated by fine-tuning an embedding model. A total of 20 top-ranked passages are then retrieved based on cosine similarity scores. To enhance retrieval performance further, we integrate BM25, a classical sparse lexical retrieval method, as a reranking tool. Passages retrieved using fine-tuned embedding model are re-ranked by combining their dense semantic scores with sparse relevance scores generated by BM25 using a weighted aggregation approach, and the top-10 results are passed as context for answer generation.\\
The embedding model was fine-tuned on a dataset derived from ObliQA for a maximum of 10 epochs, employing a batch size of 64 and a learning rate of \(2 \times 10^{-5}\). The dataset consists of anchor-positive pairs. We used Multiple Negative Symmetric Ranking Loss (MNSR) for contrastive learning, which treats every in-batch example as a potential negative example for all other queries, maximizing efficiency during training. The "symmetric" aspect means it considers bidirectional relationships (query-to-passage and passage-to-query) to improve the alignment of representations. The dev dataset was used for creating the evaluation dataset for fine-tuning, in order to load the best checkpoint at the end of the training. The model fine-tuned under this approach includes BGE-small-en-v1.5 \cite{bge_embedding}, Contextual Document Embeddings (CDE) Small \cite{morris2024contextualdocumentembeddings}, MedEmbed \cite{balachandran2024medembed}, MPNet V2 \cite{song2020mpnetmaskedpermutedpretraining}, and Stella 400M English \cite{stella2024}. The best model is used for retrieving relevant passages using LeSeR approach.\\
For the answer generation task, we test four open-source models, namely Qwen2.5 7B \cite{qwen2.5}, Mistral 7B \cite{jiang2023mistral7b}, Mistral Nemo 12B \cite{mistral_nemo}, and Gemma-2 9B \cite{gemmateam2024gemmaopenmodelsbased}. The prompts for answer generation models are designed to incorporate the retrieved passages in Sub-task 1 as contexts and the inference is done using batch size of 1. For faster inference, we use Unsloth's FastLanguageModel \cite{unsloth2024} for 2x inference performance. For assessing the performance of answer generation, we use RePASs metric \cite{gokhan2024regnlpactionfacilitatingcompliance} which measures the overall quality of answer generation using query, retrieved passages and answer, based on Entailment and Contradictions scores.\\
For assessing the performance of retrievals, we used Recall@10, which measures the proportion of relevant passages retrieved within the top-10 results, and mean Average Precision@10 (mAP@10), which evaluates the precision of ranked passages.

\section{Results}
\begin{table}[]
\centering
\begin{tabular}{lcc}
\hline
\textbf{Model}  & \textbf{Recall@10} & \textbf{mAP@10} \\ \hline
BM25 (baseline) & 0.7611             & 0.6237          \\ \hline
MPNet           & 0.6897             & 0.0949          \\
CDE             & 0.1012             & 0.0232          \\
MedEmbed        & 0.6830             & 0.0938          \\
Stella          & 0.7756             & 0.1036          \\
BGE             & 0.7040             & 0.0960          \\ \hline
MPNet\_MNSR      & 0.7977             & 0.1081          \\
CDE\_MNSR        & 0.7030             & 0.1029          \\
MedEmbed\_MNSR   & 0.8049             & 0.1108          \\
Stella\_MNSR     & 0.7973             & 0.1089          \\
BGE\_MNSR        & 0.8068             & 0.1077          \\ \hline
BGE\_LeSeR       & \textbf{0.8201}    & \textbf{0.6655} \\ \hline
\end{tabular}
\caption{Results of the retrieval task on the test dataset. Models with '\_MNSR' reprsent fine-tuned versions of the model and '\_LeSeR' represents retrieval with LeSeR approach.}
\label{tab:test1}
\end{table}

\begin{table*}[]
\centering
\begin{tabular}{lcccc}
\hline
\textbf{Method}        & \textbf{E}      & \textbf{C}      & \textbf{OC}     & \textbf{RePASs} \\ \hline
BGE\_LeSeR + Mistral 7B & 0.5229 & 0.5408 & 0.0329 & 0.3383 \\
BGE\_LeSeR + Nemo 12B   & 0.4283 & 0.4804 & 0.0353 & 0.3277 \\
BGE\_LeSeR + Gemma-2 9B & 0.5407 & \textbf{0.3262} & 0.0678 & 0.4274 \\
BGE\_LeSeR + Qwen2.5 7B & \textbf{0.5730} & 0.3480 & \textbf{0.0772} & \textbf{0.4340} \\ \hline
\end{tabular}
\caption{Results of answer generation task using RePASs on the unseen questions set. E, C, OC, and RePASs represent Entailment, Contradiction, Obligation Coverage and RePAS score, respectively.}
\label{tab:test2}
\end{table*}

During the fine-tuning phase, various retrieval models were assessed on the test dataset to identify the top-performing systems for the retrieval task (Table \ref{tab:test1}). The baseline BM25 model achieved a Recall@10 of \(0.7611\) and mAP@10 of \(0.6237\), setting a strong benchmark for comparison. Among the other models, Stella achieved a Recall@10 of \(0.7756\) and mAP@10 of \(0.1036\), demonstrating its strong retrieval performance, however poor ranking performance. Additionally, BGE reached a Recall@10 of \(0.7040\) and mAP@10 of \(0.0960\). The dense search models performed very poorly in terms of average precision compared to the baseline lexical model, suggesting that exact keyword matching might be more appropriate for the tasks.\\
Fine-tuning the dense model improved their performance in recall significantly. BGE\_MNSR (fine-tuned with MNSR loss) performed the best with Recall@10 of \(0.8068\), outperforming the baselines model. MedEmbed model, which itself is a fine-tuned version of BGE, performed similar to BGE with recall@10 of \(0.8049\). However, the semantic retrieval models still lagged behind baseline BM25 in terms of mAP@10 massively, with MedEmbed having the best mAP@10 of \(0.1108\), compared to baseline mAP@10 of \(0.6237\).
Because of its highest recall score, BGE\_MNSR is implemented in LeSeR approach. Its performance improved massively compared to its previous counterparts. With recall@10 of \(0.8201\) and mAP@10 of \(0.6655\), it outperforms all other models, including the baseline model. This shows the effectiveness of LeSeR approach in regulatory retrieval systems.\\
For assessing the performance of answer generation task, we use the answers generated by model for unseen questions, giving account of real world performance of the system. The performance of models integrating the BGE\_LeSeR retrieval system with various large language models (LLMs) was assessed using the RePASs metric and is shown in Table \ref{tab:test2}. Among the models tested, Qwen2.5 7B outperformed the others across all metrics, achieving the highest score for Entailment (\(0.5730\)), second lowest Contradiction score (\(0.3480\)), highest Obligation Coverage (\(0.0772\)), and highest RePASs (\(0.4340\)). These results demonstrate Qwen2.5's effectiveness in generating high-quality answers, making it the top performer in this evaluation. Gemma-2 9B came close to the performance of Qwen model with RePASs score of \(0.4274\) and had the lowest Contradiction score of \(0.3262\). Mistral 7B, and Nemo 12B showed comparatively lower performance across the board, with Qwen2.5 consistently outperforming them.

\section{Conclusion}

Our results highlight the significant impact of leveraging hybrid approaches to improve performance in complex retrieval and answer generation tasks. The BGE\_LeSeR when paired with Qwen2.5 7B, demonstrated superior performance in both recall and answer quality, outperforming other models such as Mistral 7B, Nemo 12B, and Gemma-2 9B across multiple metrics. Our LeSeR approach demonstrated significant improvements in both recall and precision of retrievals. This progression from traditional retrieval models to advanced LLM-based fine-tuning with reranking illustrates the importance of iterative adaptation, allowing models to specialize in retrieving relevant information while simultaneously enhancing their ability to generate coherent, contextually relevant answers.

The superior performance of Qwen2.5, particularly in the RePASs evaluation, underscores the potential of integrating fine-tuned retrieval systems with high-performing generative models to address nuanced tasks such as answer synthesis. This work emphasizes the importance of combining robust retrieval mechanisms with effective answer generation strategies to create AI systems capable of delivering high-quality, actionable insights. Integration of sophisticated embeddings and large-scale language models within the LeSeR framework demonstrates the transformative potential in improving compliance monitoring and regulatory interpretation workflows. Moving forward, future research could explore advanced fine-tuning techniques, ensemble models, newer reranking mechanisms, and domain-specific adaptations to further enhance the scalability and interpretability of these systems in regulatory domains.

\section*{Limitations}
The proposed framework, while demonstrating significant advantages, has certain limitations that should be considered. First, while dense retrieval models such as BGE\_MNSR showed substantial improvements in recall after fine-tuning, they underperformed in ranking precision, as evidenced by lower mAP@10 scores compared to the baseline BM25 model. This indicates a challenge in effectively prioritizing the most relevant passages, which is critical for practical applications requiring precise rankings. Second, dense retrieval models exhibited limitations in capturing fine-grained semantic nuances compared to lexical-based models like BM25. This shortfall may stem from the complex and diverse terminology characteristic of regulatory texts, where exact keyword matches often play a critical role. Finally, metrics such as Recall@10 and mAP@10 evaluate different aspects of retrieval performance. Recall@10 emphasizes the breadth of retrieval but does not reflect the relevance or ranking order of retrieved passages as effectively as mAP@10. The divergence in these metrics underscores the trade-offs between recall-oriented and precision-oriented evaluations, complicating the interpretation of model effectiveness. Future research should explore hybrid retrieval methods, optimize semantic understanding in dense models, and refine evaluation metrics to balance recall and precision more effectively. The answer generation could also be improved further by appending only relevant contexts in the input prompt, instead of top-10 or top-20 retrievals.

\bibliography{custom}




\end{document}